# Assessing Performance of Aerobic Routines using Background Subtraction and Intersected Image Region

*Faustine* John[1], *Irwandi* Hipiny[1,*], *Hamimah* Ujir[1] and *Mohd Shahrizal* Sunar[2]

[1]Faculty of Computer Science, Universiti Malaysia Sarawak, 94300 Kota Samarahan, Sarawak, Malaysia
[2]School of Computing, Faculty of Engineering, Universiti Teknologi Malaysia, Johor, Malaysia.

**Abstract.** It is recommended for a novice to engage a trained and experienced person (i.e., a coach) before starting an unfamiliar aerobic or weight routine. The coach's task is to provide real-time feedbacks to ensure that the routine is performed in a correct manner. This greatly reduces the risk of injury and maximise physical gains. We present a simple image similarity measure based on intersected image region to assess a subject's performance of an aerobic routine. The method is implemented inside an Augmented Reality (AR) desktop app that employs a single RGB camera to capture still images of the subject as he or she progresses through the routine. The background-subtracted body pose image is compared against the exemplar body pose image (i.e., AR template) at specific intervals. Based on a limited dataset, our pose matching function is reported to have an accuracy of 93.67%.

## 1 Introduction

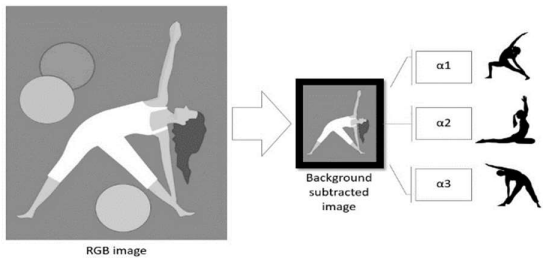

**Fig. 1.** Our proposed framework.

According to [1], the worldwide revenue in fitness segment encompassing two product categories: portable, connected fitness devices (i.e., wearables) and digital fitness and nutrition apps is amounting to US$16,221 millions in 2018. With the growing interest in personal fitness and the budding trend of exercising anytime and anywhere, there is a need for a software-based solution to measure user's performance. Ideally, the measurement has to be marker-less since the use of multiple body-worn sensors is cumbersome and costly. This solution replaces the role of a human coach in both assessing and guiding the subject as he or she performs an aerobic or weight routine. It is vital for the routine to be performed correctly to prevent and reduce the risk of injury. Moreover, by maintaining a correct posture throughout the routine greatly increases its effectiveness.

We use a simple image similarity measure, implemented inside a desktop AR app, to measure a subject's performance of an aerobic routine. The desktop app uses a single RGB camera as the solitary input to capture the subject's body pose at specific intervals. The exemplar (body pose) image, i.e., the AR template, is then superimposed over the subject's background-subtracted body pose image. The ratio of the intersected region to the union region determines the similarity score. See Figure 1 for a visualisation of our proposed framework.

Computer vision approaches have been used to recognise human activities, e.g., [2-5], human body pose and motion [34], and measuring task performance, e.g., [6] and [7]. Models are either learned from image or motion-based features. We adopt a similar approach in matching the subject's body pose to known exemplars by measuring the image similarity. The similarity score is then used to determine how well the subject had matched the required body posture at different stages of the aerobic routine.

## 2 Related Work

### 2.1. Fitness mobile apps

The following mobile apps use static videos/animated gifs to demonstrate an aerobic or weight routine to the user. No measurement, except for time-to-completion, is made since these apps are not paired with any sensors. Runtastic Result [8] is an app available on both Android and iOS platform. The app provides a series of workouts for the user aiming to either lose weight, build muscle mass or improve stamina. The app provides purchasable weekly training plans. The workouts are in the form of downloadable videos. Each video contains a countdown timer and a voice feature to alert the user to the start and end of an exercise. Next, KINETICOACH [9] is an app offered exclusively for iOS. Similar to [8], the app allows the user to customise a workout plan using a custom workout builder. The KINETICOACH app is capable of automatically suggesting the most suitable workout (for the user) based on the set goals and declared experience level during account setup.

* Corresponding author: mhihipni@unimas.my

The following apps/software are paired with wearables or devices equipped with multiple sensors such as gyroscope, accelerometer, GPS and depth camera. Fitbit Coach [10] is a personalised training app made for Android and iOS. Fitbit Coach provides workout guides in the form of video and audio instructions. The workout guides are customisable based on the set goals and current capabilities of the user. The app is actually a companion app of a wearable fitness device, i.e., Fitbit [11]. Based on the data captured by Fitbit, the app recommends a customised exercise plan to fit the user's body requirements. Another example is Nike+ Kinect Training [12], which is a video game for Xbox 360. The game requires an RGB-D sensor (i.e., Kinect camera). The Kinect camera captures a 3D point cloud representing the user body pose. The 3D point cloud is compared against the template to determine the game score.

### 2.2 Image similarity measures

For cases where the camera's pose and distance from the subject is approximately uniform during training and testing, a simple image similarity measurement such as Bobick and Davis's [13] temporal templates shall suffice. Their method performs background subtraction and aggregate the temporal foreground blobs into a single image (i.e., template). For situations where the above assumption is invalid, similarity between images can be determined by counting invariant key point matches, for example using Lowe's [14] Scale-invariant feature transform (SIFT). SIFT builds oriented gradient histograms at scale extrema, making it very robust to scale and orientation changes. Another option is to use dense feature descriptor such as Dalal and Triggs's [15] Histogram of Oriented Gradients (HOG). The higher dimensionality gives HOG improved accuracy over SIFT at the cost of reduced robustness over scale changes.

### 2.3 Background subtraction

In the initial work [16] on background subtraction, difference-of-images from television footages were computed and analysed to derive knowledge on stationary and mobile objects inside a scene. Afterwards, a significant body of works had appeared to further refine the segmentation result using the method proposed in [16]. Succeeding algorithms operate by marking pixels that does not fit the learned foreground models. Examples are pixel-wise modelling and prediction based on colour and gradient information [17], Kalman Filter [18, 19] and Gaussian Mixture Models (GMM) [20, 21]. Segmentation accuracy was further improved by exploiting known shapes and appearances of foreground objects [22-24]. These methods simplified the problem by assuming a static background. Nevertheless, these methods conveniently ignored dynamic textures such as swaying trees and moving crowds. Such structured dynamism inside the background is tackled with the use of auto-regressive models [25, 26] and kernel density estimation [27].

A novel pixel-wise and non-parametric background subtraction method was proposed in [28]. The method uses both spatial and temporal features to segment moving objects. In [29], sample background values at each pixel are quantised into a codebook. The codebook represents a compressed form of a background model for a long image sequence. This enables the capturing of structural background variations due to periodic-like motion over a long period of time. In [30], a pre-trained convolutional network is used to embed an image at multiple scale levels inside the feature space. It then uses a transposed convolutional network to learn a mapping from feature space to image space. This produces a foreground segmentation probability mask for a variety of challenging scenes such as illumination changes, background or camera motion, camouflage effect, shadow et cetera.

## 3 Our Method

### 3.1 AR templates

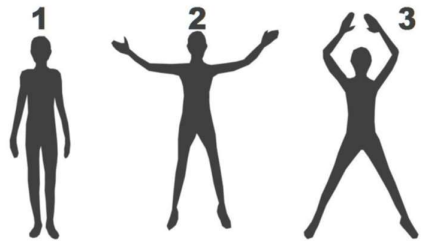

**Fig. 2.** The AR templates (stylised) for the jumping jack routine.

The AR template refers to the background-subtracted silhouette image of an expert demonstrating the correct pose at a specific moment during the aerobic routine. Figure 2 shows 3 AR templates, ordered sequentially, for the jumping jack routine. The desktop app shows the AR templates in succession (for the user to mimic) based on a pre-determined order. The user ability to mimic the pose templates determines the game score.

Table 1. The aerobic routines.

| Routine | Number of templates |
|---|---|
| Jumping jack | 3 |
| Squat | 3 |
| Lateral flexion stretches | 3 |
| Shoulder front raises | 3 |

The AR templates were manually chosen to represent the different peak moments of an aerobic routine. We selected up to 3 templates per routine. Since we have 4 aerobic routines, see Table 1, we obtained a total of 4×3=12 AR templates for the training set.

### 3.2 Pose similarity measurement

The desktop app visualises a rectangle-shaped window to guide the user to the correct position and distance from the RGB camera. Using this simplistic approach, the position and scale of the user's silhouette should be roughly identical to the AR template. Currently, the AR templates represent only one body type (i.e., we employed one expert to create all AR templates for our training set).

During the initialisation stage, the stationary camera captures the background image, (minus the user) and later subtract it from the user's body pose image. A union region is obtained by super-imposing the user's background-subtracted body pose image over the AR template. By determining the ratio of the intersected region to the union region, we obtain the similarity score, $\alpha$,

$$\alpha = \frac{\beta \cap \varepsilon}{\beta \cup \varepsilon} \quad (1)$$

where $\beta$ is the nonzero pixel region of the user's silhouette image and $\varepsilon$ is the nonzero pixel region of the AR template. Using Nearest neighbor (NN) matching, we obtain the pose classification by comparing α values of the user's silhouette image against all known AR templates.

## 4 Results and Analysis

To measure the accuracy of our pose matching function, we prepared a dataset of 12×5×6 =300 AR templates, see Table 1. A total of 5 adult participants, performed the required body pose for every AR template in six repetitions each. The participants attempted to perform the required body pose correctly for the first 3 tries, and incorrectly (on purpose) for the last 3 tries. Sample outputs from the AR desktop app are shown in Figure 3.

For each body pose image, the top-matched AR template using NN returns a similarity score between 0.0 to 1.0. By varying the threshold value at an interval of 0.1, we obtained the ROC curve in Figure 4, where

$$sensitivity = \frac{TP}{TP+} \quad (2)$$

and

$$1 - specificity = 1.0 - \frac{TN}{TN+FP} \quad (3)$$

are calculated for each threshold value.

Based on the ROC curve, the optimal threshold value is 0.8. Lower threshold values have a higher sensitivity but lower specificity. Our goal is to obtain a high true positive rate whilst maintaining a low false positive rate. With the threshold value set at 0.8, we obtained $\frac{149}{150}$ True Positives and $\frac{132}{150}$ True Negatives. The confusion matrix is shown in Figure 5. Thus, the accuracy, $ACC = 93.67\%$, is calculated using,

$$ACC = \frac{TP+TN}{TP+FP+TN+F} \quad (4)$$

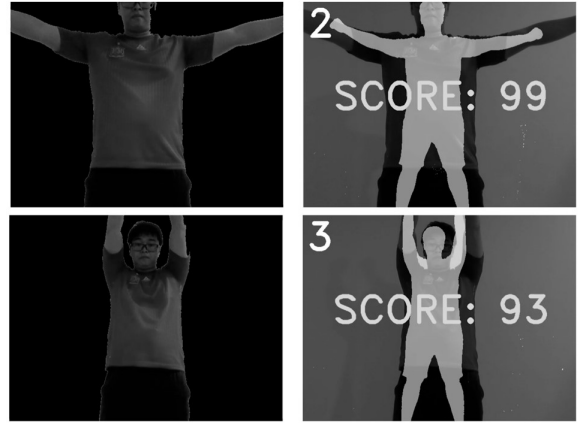

**Fig. 3.** Two sample output pairs with similarity score (over 100). Top row: 2nd AR template, and bottom row: 3rd AR template, belonging to the Jumping jack routine.

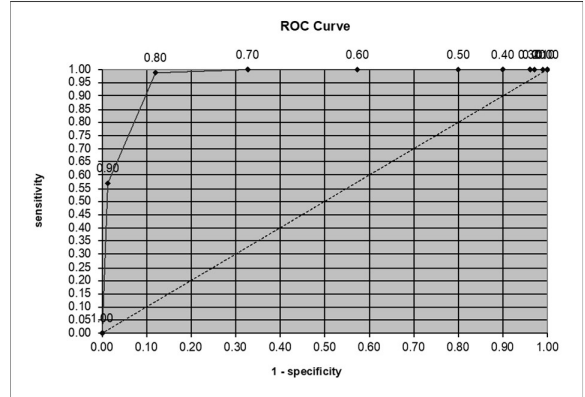

**Fig. 4.** The ROC curve. The optimal threshold value is 0.8.

|  | **Predicted Body Pose** | |
|---|---|---|
| **Actual Body Pose** | True Positives (TP) = 149 | False Positives (FP) = 18 |
| | True Negatives (TN) = 132 | False Negatives (FN) = 1 |

**Fig. 5.** The confusion matrix obtained using the optimal threshold value of 0.8.

## 5 Conclusion and Future Work

Our proposed method has a pose matching accuracy of 93.67%. The accuracy could be improved further by dynamically-resizing the AR template to match the subject's body type. Also, by using a more sophisticated image descriptor such as SIFT [14] or HOG [15] instead of intersected pixel region.

For future work, we have several plans to improve the usability of the desktop AR app. Firstly, in order to add a new aerobic routine, we have to manually segment the video into (a maximum of) 3 parts. An automated segmentation of the input video based on the discovery of natural cuts [31] would be ideal. Secondly, a measure of facial expression intensity [32] could be implemented in order to moderate the workout's intensity. If the subject displays an intense pain facial expression whilst exercising, the app may decrease the workout's intensity accordingly. The desktop app could also employ facial recognition [33] to provide customised training plan for individuals.

This work was supported by the Malaysian Ministry of Education through the following RU-RACE grant: RACE/c(2)/1253/2015(09).